\documentclass[a4paper, 10pt, conference]{ieeeconf}      
\usepackage{FG2021}
\usepackage{graphics} 
\usepackage{epsfig} 
\usepackage{mathptmx} 
\usepackage{times} 
\usepackage{amsmath} 
\usepackage{makecell}
\usepackage[T1]{fontenc}
\usepackage{bbding}
\usepackage{wasysym}
\usepackage{amssymb}
\usepackage{colortbl}
\usepackage{mathrsfs}
\usepackage{multirow}
\usepackage{footmisc}
\usepackage{tablefootnote}
\usepackage[accsupp]{axessibility}

\usepackage{amssymb}
\usepackage{pifont}
\newcommand{\cmark}{\ding{51}}%
\newcommand{\xmark}{\ding{55}}%
\def\FGPaperID{260} 
\usepackage{balance}
\newcommand*{\red}{\textcolor{red}}
\newcommand{\eg}[1]{\textit{e.g.} }
\newcommand{\ie}[1]{\textit{i.e.} }
\newcommand{\etal}[1]{\textit{et al.} }
\newcommand{\mypar}[1]{\vspace{0.3cm}\noindent\textbf{#1}}

\newcommand\footnoteref[1]{\protected@xdef\@thefnmark{\ref{#1}}\@footnotemark}

\usepackage{footnote}
\makesavenoteenv{tabular}

\usepackage{booktabs}
\title{\LARGE \bf
Affect-DML: Context-Aware One-Shot Recognition of Human Affect using Deep Metric Learning
}

\usepackage{xcolor}

\usepackage[utf8]{inputenc}

\usepackage[mathscr]{euscript}
\DeclareMathOperator*{\argmax}{\arg\!\max}

\newcommand\doubleplus{+\kern-0.5ex+\kern0.8ex}

\makeatletter
\let\NAT@parse\undefined
\makeatother

\usepackage[colorlinks]{hyperref}

\FGfinalcopy

\IEEEoverridecommandlockouts                              

\def\FGPaperID{260} 

\title{\LARGE \bf
Affect-DML: Context-Aware One-Shot Recognition of Human Affect using Deep Metric Learning
}

\author{\parbox{16cm}{\centering
    {\large  Kunyu Peng \quad \quad  Alina Roitberg \quad \quad David Schneider \\ \vspace{0.2cm} 
    Marios Koulakis  \quad \quad \quad
    Kailun Yang  \quad \quad \quad Rainer Stiefelhagen}\\
    {\normalsize
   \normalsize
     \vspace{0.3cm} Institute for Anthropomatics and Robotics, Karlsruhe Institute of Technology, Germany
    }
    {\\ \tt\small \{firstname.lastname\}@kit.edu}
    }
}

\begin{document}

\ifFGfinal
\thispagestyle{empty}
\pagestyle{empty}
\else
\author{Anonymous FG2021 submission\\ Paper ID \FGPaperID}
\pagestyle{plain}
\fi
\maketitle

\begin{abstract}
Human affect recognition is a well-established research area with numerous applications, \eg, in psychological care, but existing methods assume that all emotions-of-interest are given a priori as annotated training examples.
However, the rising granularity and refinements of the human emotional spectrum through novel psychological theories and the increased consideration of \emph{emotions in context} brings considerable pressure to data collection and labeling work. 
In this paper, we conceptualize \textit{one-shot recognition of emotions in context} -- a new problem aimed at recognizing human affect states in finer particle level from a single support sample.
To address this challenging task, we follow the deep metric learning paradigm and introduce a multi-modal emotion embedding approach which minimizes the distance of the same-emotion embeddings by leveraging complementary information of human appearance and the semantic scene context obtained through a semantic segmentation network.
All streams of our context-aware model are optimized jointly using weighted triplet loss and weighted cross entropy loss.
We conduct thorough experiments on both, categorical and numerical emotion recognition tasks of the Emotic dataset adapted to our one-shot recognition problem, revealing that categorizing human affect from a single example is a hard task. 
Still, all variants of our model clearly outperform the random baseline, while leveraging the semantic scene context consistently improves the learnt representations, setting state-of-the-art results in one-shot emotion recognition. 
To foster research of more universal representations of human affect states, we will make our benchmark and models publicly available to the community under \url{https://github.com/KPeng9510/Affect-DML}. 
\end{abstract}

\section{INTRODUCTION}
\textit{Can one recognize human emotional state given a single visual example?}
Human affect, also referred as human emotion, has always been a great concern of psychology research, as it is directly linked to the mental- and  physical health~\cite{emotion_healthy}.
Assessment of human emotion allows doctors to diagnose conditions such as Parkinson~\cite{facial_parkinson_disease}, Huntington~\cite{facial_huitingtong_disease}, Depression~\cite{facial_depression_disease} and Alzheimer~\cite{facial_alzheimer_disease} early on.
Positive emotions have capability to make human lives better and lead to better health and more productive work~\cite{emotion_review_physical}.
\begin{figure}[t]
\begin{center}
\includegraphics[width=0.95\linewidth]{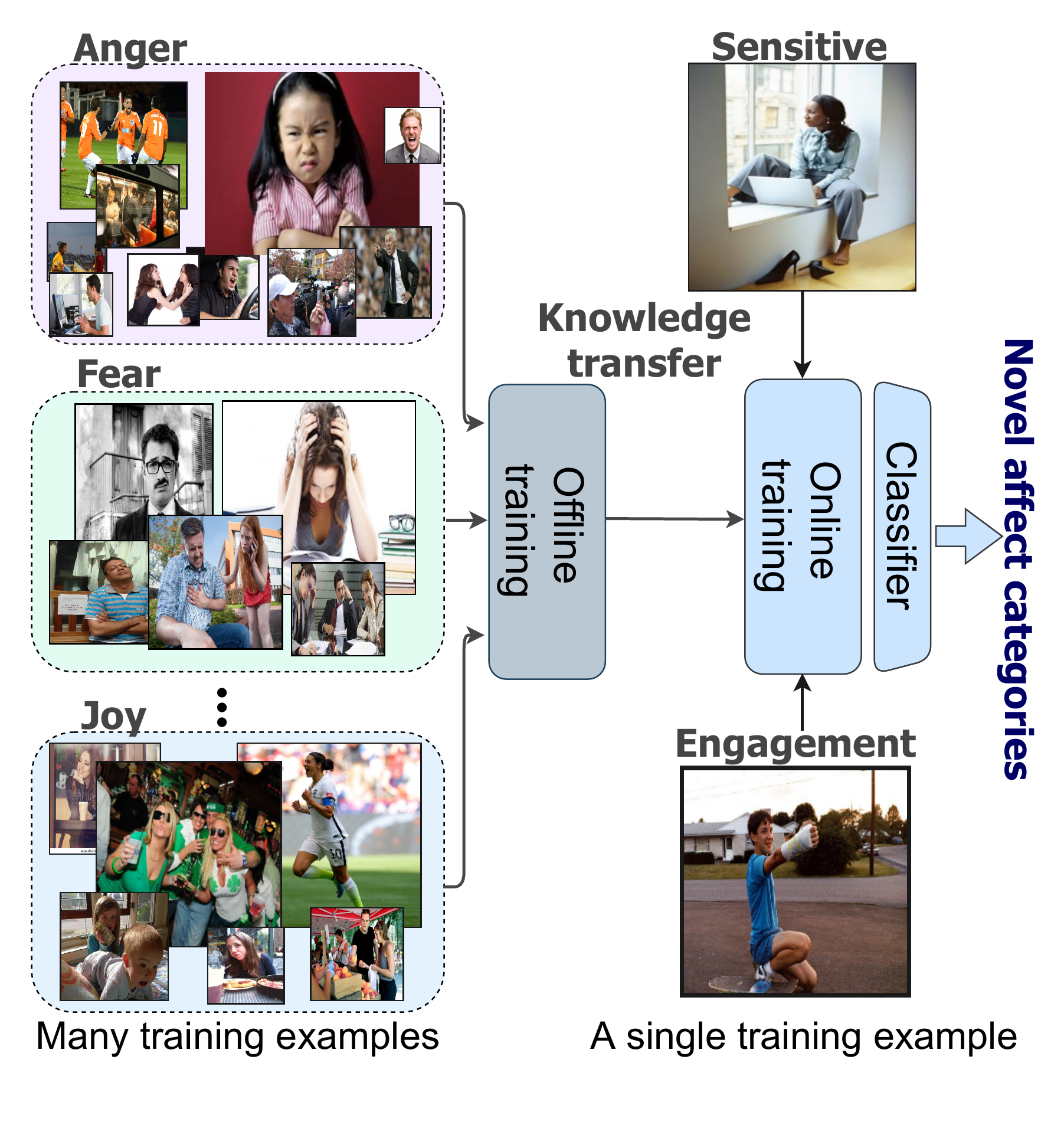}
\end{center}
\vspace{-0.9cm}
\caption{An overview of the  \textit{one-shot recognition of emotions in context} task. The a priori knowledge acquired from a label-rich dataset of known categories is transferred to categorize human affect in finer classes given a single visual example. Such one-shot recognition is a hard but important task, as it detaches changes in context-of-interest or novel psychological categorization schemes from costly data collection and labelling.}
\label{fig:statement}
\vspace{-0.5cm}
\end{figure}

Deep learning has lead to great success in many human observation tasks~\cite{carreira2017quo, liu2017skeleton, masi2018deep, roitberg2020cnn}, and also strongly influenced the field of visual emotion recognition~\cite{cnn_lstm_emotion,affectnet,emotion_efficientnet,emotic_dataset, emotic_dataset_2}.
However, deep Convolutional Neural Networks (CNNs) are known to be data-hungry and fairly limited in their ability to adapt to new situations. 
Like in many other computer vision fields, the vast majority of human affect recognition research assumes that a high amount of labelled examples is available during training for all flavours of emotions we want to recognize~\cite{cnn_lstm_emotion,affectnet,emotion_efficientnet,emotic_dataset}.
This is problematic for two reasons. 
First, human affect is a complex topic and the categorization schemes constantly evolve and become increasingly refined following the progress of psychological theories~\cite{psychological_development, ekman1999basic}. 
For example Cowen \etal~\cite{psychological_27} identified $27$ types of human affect, where the emotions can be generalizations, specializations or combinations of each other. 
Second, the increasing interest in emotions-in-context~\cite{emotic_dataset} leads to the set of possible situations being much more diverse and dynamic, eventually changing over time, as we cannot capture and annotate all possible types of environments influencing the emotion.
Bringing the idea of categorizing new flavors of emotions without excessive data labelling to the scope of researchers’ attention is therefore the main motivation of our work.

\begin{figure*}[t]

\begin{center}
\includegraphics[width=0.95\linewidth]{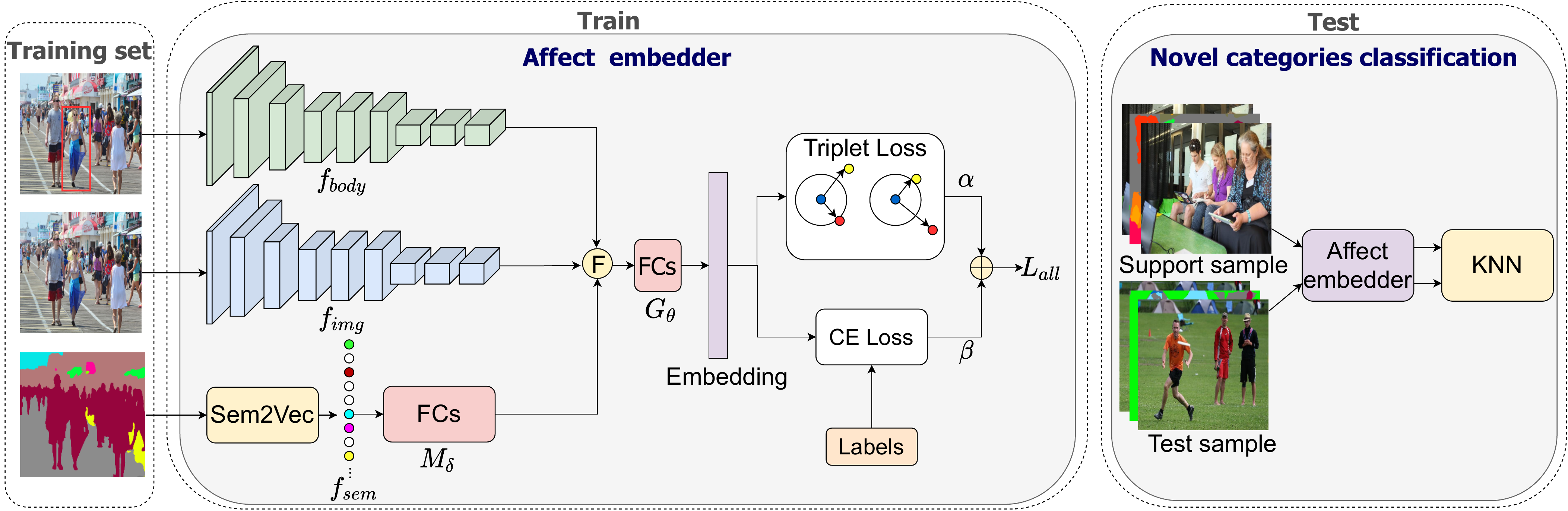}
\end{center}
\caption{An overview of the main structure for \emph{Affect-DML} proposed in this work. The input features are original full image input, human body image cropped from original input and the semantic segmentation of the full image. Three separate networks are leveraged for feature extraction. In addition to the full image feature extraction network and body feature extraction network, an MLP based semantic feature extraction network is utilized based on a embedded semantic vector calculated through leveraging semantic segmentation of the given image to provide more semantic cues for the representation extraction. Note that we rename the image feature extraction network in~\cite{emotic_dataset_2} into full image feature extraction network in our work. Triplet-margin loss and cross entropy loss with weights are leveraged for supervision of the representation extraction and K-nearest-neighbour (KNN) approach is utilized to classify the samples from unseen classes according to the distance  with the support set samples in embedding space.}
\label{fig:main}
\end{figure*}

In this work, we propose to incorporate generalization to previously unseen human affects in finer particle level through \textit{a single visual example} in the evaluation of emotion recognition models and introduce the \textit{one-shot recognition of emotions in context} task (see problem formulation in Figure~\ref{fig:statement}).
Given a label-rich dataset of \textit{known} categories, we aim to train a model which learns discriminative emotion embeddings that are well-adaptable for classification of \textit{novel} label-scarce categories.
To meet the challenge of learning generalizable affect embeddings, we follow the deep metric learning~\cite{sl-dml} paradigm, augmenting the existing emotion classification models~\cite{emotic_dataset_2} trained on Emotic dataset~\cite{emotic_dataset} with additional triplet loss~\cite{triplet_margin_loss}, which minimizes the distance of the same-emotion embeddings. 
Furthermore, we introduce a new architecture for one-shot emotion classification, which leverages the \textit{semantic scene context} obtained from a semantic segmentation framework. 
We believe that our \textit{Sem2Vec} context embeddings carry highly discriminative information specifically for emotions \textit{in context}, as it allows to effectively interpret the surroundings in the case of scarce training data, which is empirically validated through our experiments.

\mypar{Contributions and Summary.}
Given the dynamic nature of environments influencing human affect and the continuous refinement of the emotion taxonomies through new findings of psychological research, we argue that reducing the burden of data collection for incoming categories of finer human-affect categories is an important and under-researched problem.
This paper makes a step towards \textit{generalizable} emotion recognition models which are able to adapt to novel categories with a single visual example and has two major contributions.
(1) First, we formalize the under-researched task of \textit{one-shot recognition of emotions in context} (see Figure~\ref{fig:statement}), extending the conventional setup~\cite{emotic_dataset} with novel label-scarce types of affect present at test time.
To this intent, we augment the original Emotic dataset~\cite{emotic_dataset} with single-example emotions present during evaluation, defining the benchmark protocols for one-shot categorization of human affect as both, categorical- and numerical (\ie,emotion assessment on a continuous scale) problems.
(2) To achieve more generalizable representations of human affect, we follow the deep metric learning paradigm and introduce an approach for context-aware one-shot human affect recognition.
We build on the off-the-shelf approach for conventional emotion classification~\cite{emotic_dataset_2}, augmenting it with a \textit{semantic scene context branch}, which obtains semantic embeddings (\textit{Sem2Vec}) of the surroundings through a semantic segmentation framework~\cite{HRNet} and is optimized jointly with the core network using triplet loss and cross entropy loss. 
Thereby, \textit{Sem2Vec} mitigates the problem of data scarcity through discriminative environment representation achieved through knowledge transfer from a segmentation model, outperforming the native affect classification architecture by $>8\%$ in our one-shot context.
To encourage future research of more \textit{universal} emotion representations and release the pressure brought by extensive labelling of new examples, we will make our benchmark available to the public.

\section{RELATED WORK}

\subsection{Human affect recognition} 
Most of the classical affect recognition methods focus on facial expressions~\cite{fabian2016emotionet,du2014compound,soleymani2015analysis} and sometimes leverage additional cues, such as body pose~\cite{mou2015group,nicolaou2011continuous,schindler2008recognizing} or eye gaze~\cite{wolf2018estimating}.
The recently emerged Emotic dataset~\cite{emotic_dataset} enabled large-scale studies of recognizing \emph{emotions in context}, considering diverse surrounding scenes influencing human affect and resulting in multiple neural architectures developed for this task~\cite{antoniadis2021exploiting,mittal2020emoticon,ruan2020context,emotic_dataset_2}.
Mittal~\textit{et al.}~\cite{mittal2020emoticon} leveraged three interpretations, \textit{i.e.}, multi-modal, attention-driven and depth-based context for dynamic emotion recognition. 
Ruan~\textit{et al.}~\cite{ruan2020context} designed an architecture that makes use of both global image context and local details of the target person for multi-label emotion recognition.
Antoniadis~\textit{et al.}~\cite{antoniadis2021exploiting} proposed to exploit emotional dependencies by using a graph convolutional network.
Differently from these works, we propose a semantic-aware one-shot human affect recognition method which handles the challenges of novel classes emerging in unconstrained conditions, while leveraging context encoding based on semantic segmentation maps. 

\subsection{One-shot recognition} 
One-shot learning has been a hot research topic over the last years. Existing methods mainly pursue the paradigms including data augmentation, transfer learning, deep embedding learning, and meta learning. Data augmentation based methods~\cite{dosovitskiy2014discriminative} enrich the training data to produce more diverse examples
or hallucinate examples of novel classes.
Transfer learning based methods~\cite{fe2003bayesian,wang2016learning} aim to reuse knowledge in learned tasks.
Deep embedding learning methods~\cite{koch2015siamese,vinyals2016matching} are built to design low-dimension embedding spaces to yield more discriminative feature representations.
Meta learning models~\cite{santoro2016meta,wang2016learning} attempt to harvest knowledge among multiple tasks.
There are additional few-shot learning methods~\cite{guo2017one,sl-dml,reiss2020activity,shaban2017one} for various tasks like face- and action recognition, driver observation, and semantic segmentation. 
The work most similar to ours is presumably the one of Cruz~\etal~\cite{cruz2014one} (2014) who proposed a Haar-like-features-based framework for one-shot classification of facial expressions.
However, the setting of~\cite{cruz2014one} is very restrictive compared to modern datasets~\cite{emotic_dataset} and puts the environment aside while focusing on frontal facial images only.
Furthermore, Wang~\textit{et al.}~\cite{wang2019robust} addressed on few-shot sentiment analysis on social media, while Zhan~\textit{et al.}~\cite{zhan2019zero} considered zero shot emotion categorization, which, however, is very different from the one-shot task as it is centered around finding good textual embeddings for the unknown emotions, while one-shot learning assumes a single visual example present at test-time.
In this work, we conceptualize a novel task of \textit{one-shot emotion recognition in context}, which has been largely overlooked in previous research.

\section{One-Shot Recognition of Human Affect}
\label{subsubsec:method:core_architecture}

\subsection{Problem Formulation \& One-Shot Emotic Benchmark}
\label{sec:problem}
We tackle the problem of \textit{one-shot recognition of emotions in context} from image data, which aims to assign the correct human affect class given a single support set sample for the unseen categories. 
First, we provide a formal definition of the problem, with an overview illustrated in Figure~\ref{fig:statement}. 
Conceptually, one-shot human affect recognition aims to transfer a priori knowledge acquired from a label-rich dataset of known categories to categorize human affect in finer-particle level given a single support set sample.

Formally, let $C_{novel}$ denote the unseen classes and $C_{bases}$ indicate the seen classes.
One-shot recognition is the task of classifying examples in a set of unseen classes $C_{novel}$ given one single sample per unseen class in the support set, and a large amount of training data for the seen classes $C_{base}$.
Formally, we start with a training set $D = \{(\mathbf{d}_n, y_n)\}_{n=1}^{N}$, $y_n\in C_{base}$.
Then a support set $R = \{\mathbf{x}_i\}_{i = 1}^{|C_{novel}|}$, where $C_{base}\cap C_{novel} = \emptyset$, and a query set $Q = \{\mathbf{q}_n\}_{n = 1}^M$ are supplied and the goal is to assign a class $y\in C_{novel}$ to each $\mathbf{q}_n\in Q$.
Note that the query set is also the testing set.

\begin{table}[t]
\centering
\caption{Human affect categorical class mapping for one-shot human affect recognition. }
\scalebox{0.95}{
\begin{tabular}{c|c|l|l}
\toprule
\multicolumn{1}{l|}{\multirow{2}{*}{Mode}} & \multirow{2}{*}{One-shot classes} &
 \multicolumn{2}{c}{Emotic~\cite{emotic_dataset} classes} \\ 
\multicolumn{1}{l|}{} &  &
 \multicolumn{1}{c|}{\emph{CAT-6:6} split} & \multicolumn{1}{c}{\emph{CAT-6:4} split} \\ 
\hline
\multirow{6}{*}{Train} & Angry & \begin{tabular}[c]{@{}l@{}}Disapproval\\Aversion \\Annoyance\\Anger\end{tabular} & \begin{tabular}[c]{@{}l@{}}Disapproval\\Aversion\\Annoyance \\Anger\end{tabular} \\ 
\cline{2-4}
 & Sadness & \begin{tabular}[c]{@{}l@{}}Suffering\\Sadness \\Fatigue \\Pain\end{tabular} & \begin{tabular}[c]{@{}l@{}}Suffering \\Sadness\\Fatigue\\Pain \\Sensitive\\Embarrassment\end{tabular} \\ 
\cline{2-4}
 & Fear & \begin{tabular}[c]{@{}l@{}}Fear \\Disquitement\end{tabular} & \begin{tabular}[c]{@{}l@{}}Fear \\Disquitement\end{tabular} \\ 
\cline{2-4}
 & Love & \begin{tabular}[c]{@{}l@{}}Peace\\Affection\\Sympathy\end{tabular} & \begin{tabular}[c]{@{}l@{}}Peace \\Affection\\Sympathy\end{tabular} \\ 
\cline{2-4}
 & Joy & \begin{tabular}[c]{@{}l@{}}Happiness \\Pleasure \\Excitement\\Anticipation\end{tabular} & \begin{tabular}[c]{@{}l@{}}Happiness \\Pleasure\\Excitement \\Anticipation\end{tabular} \\ 
\cline{2-4}
 & Surprising & \begin{tabular}[c]{@{}l@{}}Surprising \\Confuse/Doubt \\Confidence\end{tabular} & \begin{tabular}[c]{@{}l@{}}Surprising \\Confuse/Doubt\\Confidence\end{tabular} \\ 
\hline
\multirow{6}{*}{Test} & Disconnection & Disconnection & Disconnection \\ 
 & Engagement & Engagement & Engagement \\ 
 & Sensitive & Sensitive & \xmark  \\ 
 & Embarrassment & Embarrassment & \xmark \\ 
 & Esteem & Esteem & Esteem \\ 
 & Yearning & Yearning & Yearning \\
\bottomrule
\end{tabular}}
\label{tab:split}
\end{table}

\mypar{Categorical benchmark.} In order to provide a benchmark for \textit{one-shot recognition of emotions in context}, we reformulate the $26$ categories covered by the well-established Emotic dataset~\cite{emotic_dataset} developed for conventional human affect classification to our one-shot recognition task. 
To achieve this, we split the dataset category-wise in the seen classes $C_{bases}$ and unseen classes $C_{novel}$, following Table~\ref{tab:split}.
As our support set $R$ we select the first sample of the corresponding unseen category generated through random order, while the remaining categories constitute the query set $Q$ also noted as test set.
Due to the trend of making human affect taxonomies more and more fine-grained, our seen classes $C_{bases}$ comprise \textit{coarser} types of human affects, which are \textit{anger}, \textit{sadness}, \textit{joy}, \textit{love}, \textit{fear} and \textit{surprise} constructed by the emotional categories provided by Emotic dataset~\cite{emotic_dataset}.
The fine-to-coarse cluster relationship of these human affect categories follows the human affect tree structure proposed by Shaver~\textit{et al.}~\cite{shaver1987emotion}.
The test set $C_{novel}$ therefore represents more fine-grained human affects.
To honor the one-shot learning premise, we reassure disjoint training and testing categories, \ie, $C_{base}\cap C_{novel} = \emptyset$ also taking into account generalizations to make sure there is no overlapping.
We consider two kinds of train-test split for the categorical one-shot human affect recognition task, denoted as \emph{CAT-6:4} and \emph{CAT-6:6}, with the detailed overview provided in Table~\ref{tab:split}.

\mypar{Reformulated numerical benchmark.} We further reformulate numerical human affect regression task into one-shot recognition task to further verify the efficiency of our approach on the unnamed human affect categories defined by different level of the three main components, which are \textit{valence}, \textit{arousal} and \textit{dominance} in Emotic dataset~\cite{emotic_dataset} based on the work from Posner~\textit{et al.}~\cite{posner2005circumplex}, considering the discrete level annotation from $1$ to $10$ as in \cite{emotic_dataset} as category labels for each component, with the splits denoted as \emph{LEV-7:3} and \emph{LEV-6:4} for different seen-unseen proportions.

\begin{figure*}[t]
\begin{center}
\includegraphics[width=0.92\linewidth]{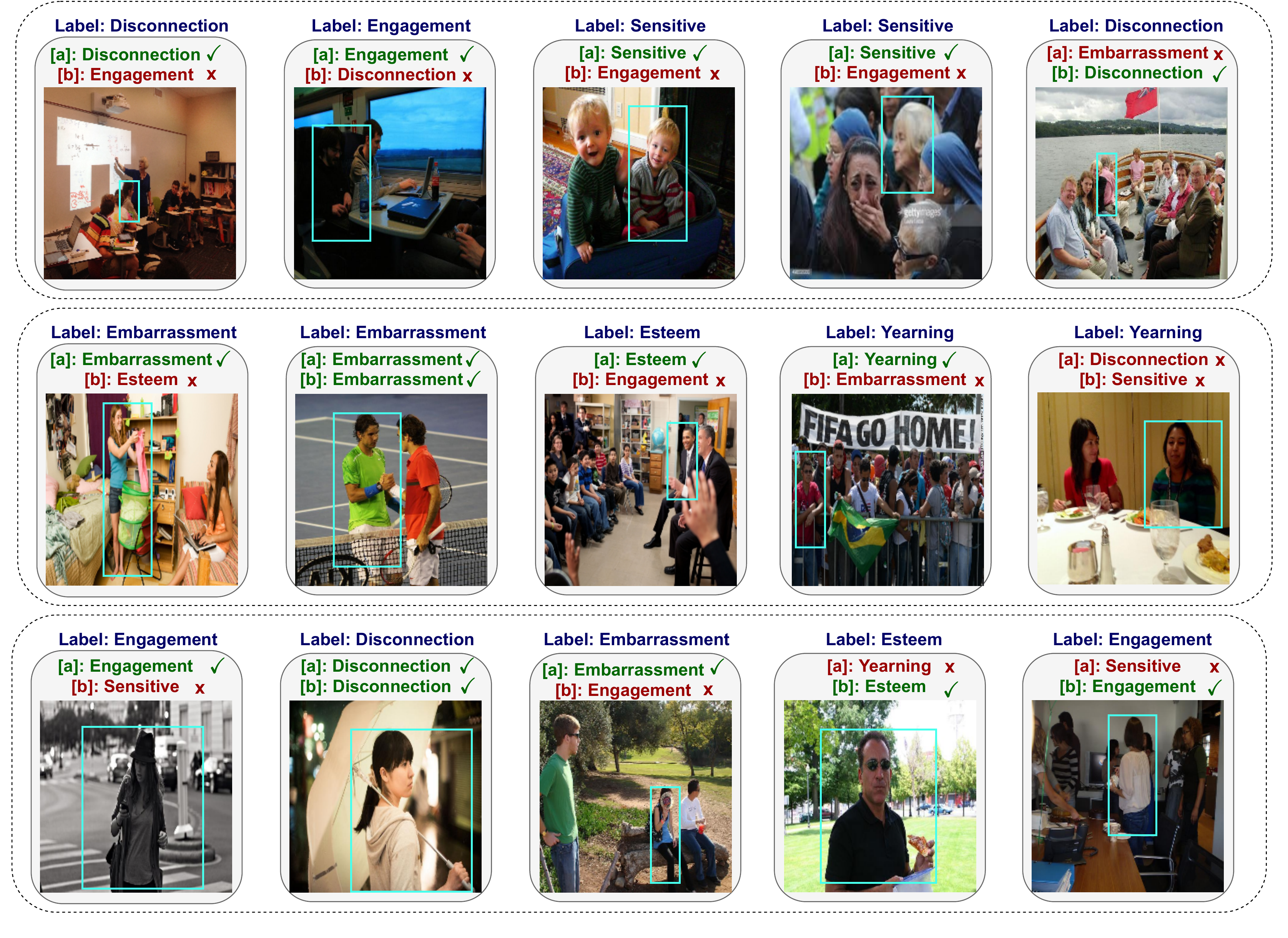}
\end{center}
\caption{A comparison of the one-shot categorical human affect recognition between our best model of \emph{Affect-DML} indicated by [a] and Kosti~\textit{et al.}~\cite{emotic_dataset_2} extended with deep metric learning (IB-DML$^{\red{\dagger}}$) indicated by [b] on the test set.
False recognition is marked as red color. The train-test split for these two models is \emph{CAT-6:6}. The blue bounding box indicates the annotated person for categorical human affect recognition according to Emotic dataset~\cite{emotic_dataset}.}
\label{fig:qualitative}
\end{figure*}
\mypar{Dealing with multi-label annotations. }Since Emotic~\cite{emotic_dataset} is a multi-label dataset, we leverage a multiple-to-single label mapping approach to realize unique label encoding for each sample, demonstrated in detail in following.
Assuming $N$ classes in total for both training and testing, let $\textbf{e}$ denote a $N$-dimensional zero vector.
If emotional labels for one sample are contained by list $W$, then the desired label $y$ for this sample can be encoded as 
\begin{equation}
\mathbf{e}[Label2index(l_{i})] ~+=~1, ~for~l_{i}~in~ W,
\end{equation}
\begin{equation}
\mathbf{y} = \argmax(\mathbf{e}),
\end{equation}
where \textit{Label2index} maps a label from Emotic dataset~\cite{emotic_dataset} into one-shot encoded label according to Table~\ref{tab:split} considering the column index for categorical human affects, and for numerical human affect one-shot recognition, $[1,3,5,6,8,9,10]$ is selected as training set for \emph{LEV-7:3} train-test split and $[1,3,5,6,8,9]$ is selected as training set for \emph{LEV-6:4} train-test split while the rest classes are set as testing set for each.
The \emph{Label2index} is identity mapping for numerical human affect one-shot recognition benchmark.
The final label for each sample is generated through an \emph{Argmax} operation according to the encoded vector $\mathbf{e}$.

\subsection{Deep Metric Learning Architecture}

To address one-shot recognition of emotions-in-context task, we follow the deep metric learning paradigm~\cite{sl-dml} (main workflow illustrated in Figure~\ref{fig:main}). 
Our model takes an image $I$ as input together with the location of the person-of-interest (as in \cite{emotic_dataset}) which we use to obtain the body crop image $I_{body}$. 
This information is passed through the networks $f_{body}, f_{img}, f_{sem}$ to produce the features $f_{body}(I_{body}), f_{img}(I_{img}), f_{sem}(I_{sem})$ capturing target person, full image and its semantic context respectively.
Those features are fused and embedded through network $f_{emb}$. 
The network is optimized to learn highly seperable human affect representations with  deep metric learning (DML).
During inference, the finer unseen  classes are classified based on the sample in support set closest to them in the embedding space.
We now describe key components of our architecture in detail.

\subsection{Feature encoders}
All three feature encoders $f_{body}, f_{img}, f_{sem}$ are off-the-shelf networks with the exception of $f_{sem}$ where we enhance an existing architecture to build a compact segmentation feature vector. 
ResNet~\cite{ResNet} classification model is leveraged to form $f_{body}$ and $f_{img}$ individually while discarding \textit{softmax} layer, and separately the crop of person $I_{body}$ and full image $I_{img}$ are utilized as input for two individual feature extraction branches.
Finally, the final semantic embedded feature can be demonstrated as $f_{sem} = M_{\delta}\circ Sem2Vec$, where $Sem2Vec$ consists of an off-the-shelf segmentation network together with a layer providing the set of predicted semantic classes in a one-hot-encoded form and $M_{\delta}$ denotes fully-connected layers. 
To elaborate on $Sem2Vec$, assume that $Sem$ is the segmentation network and $Sem(I)$ denotes the semantic segmentation result of image $I$. 
If $Set\circ Sem(I)$ is the set of all distinct classes appearing on at least one pixel, and $X = Sem2Vec(I)$, then 
\begin{equation}
X[i] = \left\{
\begin{array}{ll}
      1, & \text{if }i\in Set\circ Sem(I) \\
      0, & \text{otherwise}\\
\end{array} 
\right.
\end{equation}

The length of $X$ is fixed as the number of pre-defined total classes of semantic segmentation network.
The features of the three branches are embedded to the vector $f_{emb}(I_{img}, I_{body}, I_{sem}) = G_{\theta}\circ F(f_{body}(I_{body}), f_{img}(I_{img}), f_{sem}(I_{sem}))$, where $F$ is concatenation operation and $G_{\theta}$ denotes a fully-connected layer, mapping the concatenated feature into desired-dimensional representation.

\subsection{Deep Metric learning}

Our training procedure builds on the  deep metric learning pipeline of~\cite{sl-dml}.
On the sampling phase, a batch of samples $B = \{(\mathbf{x}_n, y_n)\}_{n=1}^{N_{b}}$ is picked and embedded via $f_{emb}$ to $B_{emb} = \{(\mathbf{z}_n, y_n)\}_{n=1}^{N_{b}}$. The batch size is denoted as $N_{b}$.
Right after, the Multi-Similarity Miner~\cite{multi_similarity} is used to select the most informative positive and negative pairs of embeddings. 
Namely, if $\{(\mathbf{z}_{a_i}, \mathbf{z}_{p_i})\}_{i=1}^{N_{p}}$, $\{(\mathbf{z}_{a_i}, \mathbf{z}_{n_i})\}_{i=1}^{N_{n}}$ are all possible positive and negative pairs of $B_{emb}$ and $S$ is cosine similarity, where $N_p$ denotes negative sample number and $N_p$ denotes positive sample number. Then the filter processing is defined as
\begin{equation}
    S(\mathbf{z}_{a_i}, \mathbf{z}_{p_i}) > \max_{y_k \neq y_{a_i}} S(\mathbf{z}_{a_i}, \mathbf{z}_{a_k}) + \epsilon
\end{equation}
\begin{equation}
    S(\mathbf{z}_{a_i}, \mathbf{z}_{n_i}) < \min_{y_k = y_{a_i}} S(\mathbf{z}_{a_i}, \mathbf{z}_{a_k}) - \epsilon
\end{equation}
Thus we keep only the positive pairs containing relatively smaller (relative to $\epsilon$) similarity than the most similar negative pair of the anchor and the negative pairs containing a slightly higher similarity than the least similar positive pair of the anchor.

The positive and negative pairs sharing the same anchor are combined to create triplets $\{(\mathbf{z}_{a_i}, \mathbf{z}_{p_i}, \mathbf{z}_{n_i})\}_{i=1}^{N_{t}}$ which are used in computing the triplet loss with margin $\lambda$ , with $N_{t}$ denotes triplet number: 

\begin{equation}
  \mathscr{L}_{triplet} = \sum_{i=1}^{N_t}[\left\|\mathbf{z}_{a_i} - \mathbf{z}_{n_i}\right\| - \left\|\mathbf{z}_{a_i} - \mathbf{z}_{p_i}\right\| + \lambda]_+
\end{equation}

Aside from being used to compute $\mathscr{L}_{triplet}$, the embeddings $B_{emb}$ are passed through a fully-connected layer and generate a classification loss $\mathscr{L}_{class}$ (cross entropy). The total loss is then a weighted sum of the aforementioned two losses:
\begin{equation}
    \mathscr{L}_{all} = \alpha \mathscr{L}_{triplet} + \beta \mathscr{L}_{class}
\end{equation}

\subsection{Inference}
Deep metric learning pushes the embedder to learn class-divisible representations in the embedding space.
This makes it possible to reduce the one-shot recognition problem to a nearest neighbour search in the embedding space, aiming at reducing the distance inside class and increase the distance between different classes.

Let $R = \{\mathbf{x}_i\}_{i=1}^{|C|}$, where $C = C_{novel}$ is the set of novel classes, be the support set of the one-shot recognition and $\mathbf{x} = (I_{body}, I_{image}, I_{sem})$ is a single example to classify. 
Also assume $R_{emb} = \{\mathbf{z}_i\}_{i=1}^{|C|}$ and $\mathbf{z}$ are the embedded vectors of those examples via $f_{emb}$. 
We make the assumption that for each class $i$, the probability of a sample belonging to that class is given by an isotropical distribution centered around $\mathbf{z}_i$. 
We also assume that the density of the distribution is strictly decreasing on the distance from $\mathbf{z}_i$, thus the density is defined as $p(\mathbf{z}|y_i) = f(\left\|\mathbf{z} - \mathbf{z}_i\right\|)$.
A concrete candidate for $f$ could be for example the density of a multidimensional isotropical normal distribution centered around $\mathbf{z}_i$. 
We further assume that the support classes are independent from each other and they occur with equal frequencies. Then by Bayes rule we have:
\begin{align}
    P(y_i|\mathbf{z}) &= \frac{p(\mathbf{z}|y_i) P(y_i)}{\sum\limits_{j\in C} p(\mathbf{z}|y_j) P(y_j)} \\
    &= \frac{p(\mathbf{z}|y_i) \frac{1}{|C|}}{\sum\limits_{j\in C} p(\mathbf{z}|y_j) \frac{1}{|C|}} = \gamma f(\left\|\mathbf{z} - \mathbf{z}_i\right\|),
\end{align}
where $\gamma$ is constant for different classes and a fixed value of $\mathbf{z}$. But then, the predicted class of $\mathbf{x}$ reduces to 
\begin{equation}
y_{pred} = \argmax\limits_{y_i\in C} f(\left\|\mathbf{z} - \mathbf{z}_i\right\|),
\end{equation}
which is equivalent to selecting the nearest neighbour of $\mathbf{z}$ in the embedding space.
Note that though the above assumptions are quite restricting, they are in some sense the best one can hope for given the very limited information one-shot learning provides us with, illustrating the ability of deep metric learning to create a good embedding space for both seen and unseen classes which decides how well our proposed approach can separate the unseen classes.

\mypar{Segmentation-driven context encoding.}
\label{subsubsec:method:semantic-encoding}
In order to extract semantic related attributes-based features from full image, \textit{Sem2Vec} approach is proposed in our work to leverage underlying semantic cues to enhance the recognition of human affect.
First, as aforementioned semantic segmentation image is generated through well-trained HRNet~\cite{HRNet} model based on ADE20K~\cite{ade20k} dataset with $N_{sem}$ semantic segmentation classes given the full image as input (we use $150$ segmentation categories in our experiments).
After the generation of binary encoded vector as aforementioned to describe which kind of object occurred in the full image and provide semantic cues for human affect recognition illustrated as $\textbf{X}=Sem2Vec(\textbf{x})$, the final representation $f_{sem}$ harvesting semantic cues is extracted based on the vector representation of the semantic information:
\begin{equation}
  f_{sem}=M_{\delta}(Sem2Vec(\textbf{x})), 
\end{equation}
where $M_{\delta}$ denotes fully-connected layers with \textit{ReLU} and normalization.

\section{EXPERIMENTS}

We conduct extensive experiments on the One-Shot-Emotic benchmark.
First, we describe the implementation details of our proposed approach in Section~\ref{subsec:experiments:implementation}.
Then, we present our quantitative results on both categorical and numerical one-shot human affect recognition in Section~\ref{subsec:experiments:human_affect_categories} and Section~\ref{subsec:experiments:human_affect_levels} accordingly.
Finally, Section~\ref{subsec:experiments:qualitative} conducts an ablation of the model architecture choices and provides qualitative examples of the predictions.

\subsection{Implementation Details}
\label{subsec:experiments:implementation}

\begin{table}[t]

\caption{Results for One-Shot Categorical Recognition.}
\centering
\label{tab:experiments_category}
\scalebox{0.95}{
\begin{tabular}{lccccc} 
\toprule
\multicolumn{1}{c}{\multirow{2}{*}{Method}} & \multicolumn{3}{c}{Setting} & \multicolumn{2}{c}{Accuracy} \\ 
\multicolumn{1}{c}{} & Image & Body & Semantic & \emph{CAT-6:6} & \emph{CAT-6:4} \\ 
\hline
\multicolumn{6}{c}{\cellcolor{gray!10}{\textbf{Baseline methods}}}\\ 
\hline 
Random & \xmark & \xmark & \xmark & 16.78 & 16.68 \\
Kosti~\textit{et al.}~\cite{emotic_dataset_2}$^{\red{\dagger}}$ & \cmark & \cmark & \xmark & 29.98 & 30.22 \\ 
\hline
\multicolumn{6}{c}{\cellcolor{gray!10}\textbf{Image/Body-based architectures~\cite{emotic_dataset_2} trained with DML}} \\ 
\hline

I-DML  & \cmark & \xmark & \xmark & 17.84 & 27.64 \\
B-DML  & \xmark & \cmark & \xmark & 15.41 & 27.66 \\
IB-DML  & \cmark & \cmark & \xmark & 21.02 & 28.19 \\ 
IB-DML $^{\red{\dagger}}$& \cmark & \cmark & \xmark & 24.20 & 29.46 \\
\hline
\multicolumn{6}{c}{\cellcolor{gray!10}\textbf{Semantic segmentation-enhanced architectures }} \\ 
\hline
Sem-I-DML  & \cmark & \xmark & \cmark & 32.16 & 33.48 \\
Sem-IB-DML  & \cmark & \cmark & \cmark & \textbf{\textbf{33.76}} & \textbf{36.26} \\
\bottomrule
\multicolumn{6}{l}{\parbox[t]{\linewidth}{$^{\red{\dagger}}$ Body feature extraction network is pretrained on Places-365~\cite{places365} and full image feature extraction network is pretrained on ImageNet~\cite{imagenet} to reproduce the setting of Kosti \etal~\cite{emotic_dataset_2} (all other models except random baseline are both pretrained using ImageNet~\cite{imagenet} for body feature extraction network and full image feature extraction network.)}}
\end{tabular}}
\end{table}

\begin{table*}[t]
\centering
\caption{Experimental results on One-Shot Emotic dataset for numerical human affect.}
\scalebox{0.8}{
\label{tab:experiment_level}
\begin{tabular}{lccc|cccl|cccl} 
\toprule
\multicolumn{1}{c}{\multirow{2}{*}{Name}} & \multicolumn{3}{c|}{Setting} & \multicolumn{4}{c|}{\emph{LEV-7:3} split Acc} & \multicolumn{4}{c}{\emph{LEV-6:4} split Acc} \\ 
\multicolumn{1}{c}{} & Image & Body & Semantic & Valence & Arousal & Dominance& \multicolumn{1}{c|}{Avg Acc} & Valence & Arousal & Dominance  & \multicolumn{1}{c}{Avg Acc} \\ 
\hline
\multicolumn{12}{c}{\cellcolor{gray!10}\textbf{Baseline methods}} \\ 
\hline
Random & \xmark & \xmark & \xmark & 12.23 & 6.86 & 13.26 & 10.79 & 7.62 & 11.73 & 13.28 & 10.88 \\ 
Kosti~\textit{et al.}~\cite{emotic_dataset_2}$^{\red{\dagger}}$ & \cmark & \cmark & \xmark & 36.03 & 45.01 & 45.02 & 42.02 & 28.68 & 33.65 & 37.24 & 34.78 \\ 
\hline
\multicolumn{12}{c}{\cellcolor{gray!10}\textbf{Image/Body-based architectures~\cite{emotic_dataset_2} trained with DML}} \\ 
\hline

I-DML  & \cmark & \xmark & \xmark & 46.72 & 46.89 & 55.63 & 49.75 & 30.63 & 38.31 & 63.36 & 44.10 \\
B-DML  & \xmark & \cmark & \xmark & 35.89 & 41.95 & 54.99 & 44.28 & 37.05 & 36.91 & 53.62 & 42.53 \\
IB-DML & \cmark & \cmark & \xmark & 50.75 & 46.03 & 59.16 & 51.98 & 48.96 & 38.20 & 58.69 & 48.62 \\ 
IB-DML (~\cite{emotic_dataset_2} with DML)  $^{\red{\dagger}}$ & \cmark & \cmark & \xmark & 49.00 & 43.53 & 56.23 & 49.59 & 47.33 & 35.13 & 55.99 & 46.15 \\
\hline
\multicolumn{12}{c}{\cellcolor{gray!10}\textbf{Semantic segmentation-enhanced architectures (ours)}} \\ 
\hline
Sem-I-DML (ours) & \cmark & \xmark & \cmark & \textbf{56.79} & 46.84 & 58.78 & 54.14 & 51.76 & 38.72 & \textbf{65.29} & 51.92 \\
Sem-IB-DML (ours) & \cmark & \cmark & \cmark & 52.36 & \textbf{47.19} & \textbf{64.52} & \textbf{54.89} & \textbf{58.92} & \textbf{41.37} & 62.56 & \textbf{54.28} \\
\bottomrule
\multicolumn{12}{@{}l}{\parbox[t]{\linewidth}{$^{\red{\dagger}}$ Body feature extraction network is pretrained on Places-365~\cite{places365} and full image feature extraction network is pretrained on ImageNet~\cite{imagenet} to reproduce the setting of Kosti \etal~\cite{emotic_dataset_2} (all other models except random baseline are both pretrained using ImageNet~\cite{imagenet} for body feature extraction network and full image feature extraction network.)}}
\end{tabular}}
\end{table*}

\begin{table*}
\centering
\caption{Experiments for different full image and body feature extraction network structure.}
\scalebox{0.85}{
\label{tab:experiments_ablation}
\begin{tabular}{l|c|c|c|c|c|c|c|c|c|c|c|c|c|c|c|c} 
\toprule
\begin{tabular}[c]{@{}l@{}}Image/body\\model\end{tabular} & \multicolumn{8}{c|}{ResNet18} & \multicolumn{8}{c}{ResNet50} \\ 
\hline
Split & \emph{CAT-6:6} & \emph{CAT-6:4} & \multicolumn{3}{c|}{\emph{LEV-7:3}} & \multicolumn{3}{c|}{\emph{LEV-6:4}} & \emph{CAT-6:6} & \emph{CAT-6:4} & \multicolumn{3}{c|}{\emph{LEV-7:3}} & \multicolumn{3}{c}{\emph{LEV-6:4}} \\ 
\hline
Affect mode & Cat. & \multicolumn{1}{c|}{Cat.} & Val. & Aro. & Dom. & Val. & Aro. & Dom. & Cat. & \multicolumn{1}{c|}{Cat.} & Val. & Aro. & Dom. & Val. & Aro. & Dom. \\ 
\hline
Accuracy & \multicolumn{1}{c}{33.76} & \multicolumn{1}{c}{\textbf{36.26}} & \multicolumn{1}{c}{52.34} & \multicolumn{1}{c}{47.19} & \multicolumn{1}{c}{\textbf{64.52}} & \multicolumn{1}{c}{\textbf{58.92}} & \multicolumn{1}{c}{41.37} & \textbf{62.56} & \multicolumn{1}{c}{\textbf{35.80}} & \multicolumn{1}{c}{36.24} & \multicolumn{1}{c}{\textbf{56.27}} & \multicolumn{1}{c}{\textbf{48.51}} & \multicolumn{1}{c}{64.51} & \multicolumn{1}{c}{57.69} & \multicolumn{1}{c}{\textbf{41.59}} & 62.22 \\
\bottomrule
\end{tabular}
}
\end{table*}


We use ResNet~\cite{ResNet} pre-trained on ImageNet~\cite{imagenet} as our body pose encoding network $f_{body}$ and image representation encoding network $f_{img}$.
Tables~\ref{tab:experiments_category} and~\ref{tab:experiment_level} provide results using the ResNet18 version of the backbone, while Table~\ref{tab:experiments_ablation} compares ResNet18 and ResNet50 to analyze the impact of the model capacity.
The input preprocessing for body pose encoding network covers cropping the person bounding boxes through annotations provided by~\cite{emotic_dataset_2} and rescaling the cropped image to  $256 \times 256$. 
For our emotion recognition architecture, we remove the last \emph{softmax} layer and append a single fully-connected layer to both human body and full image networks in order to project the output logits into a fixed $512$-dimensional feature space.
The aforementioned $M_{\delta}$ with channel setting $[256,512]$, leveraging a Rectified Linear Unit (ReLU) as activation function followed by a batch normalization layer between fully-connected layers with batch size $32$.
RMSprop~\cite{rmsprop} is used as optimizer with learning rate $3.5e^{-4}$ for categorical human affect experiments described in Section~\ref{subsec:experiments:human_affect_categories}. 
For the numerical human affect recognition experiments described in Section~\ref{subsec:experiments:human_affect_levels}, the learning rate is also set to $3.5e^{-4}$ for the \emph{LEV-7:3} train-test split and to $3.5e^{-6}$ for the \emph{LEV-6:4} (the learning rate was chosen using grid search to optimize the validation set performance).
The weight decay factor $\gamma$ is universally set to $0.1$ with a step size of $4$ epochs.
Our approach is trained directly with the triplet loss and cross entropy loss with weights setting $0.5$ for $\alpha$ and $0.5$ for $\beta$ to balance representation learning and classification. 
Further details are listed in the supplementary.
\subsection{One-shot Categorical Recognition Results}
\label{subsec:experiments:human_affect_categories}
Table~\ref{tab:experiment_level} compares the original approach of Kosti~\textit{et al.} ~\cite{emotic_dataset_2} developed for standard emotion recognition with different variants of our enhancement with DML training and our proposed segmentation-aware model.
Different split types are marked as \emph{CAT-6:6} and \emph{CAT-6:4} (see Section \ref{sec:problem} for details).
We also consider the impact of the individual branches, \textit{i.e.}, the image- (\emph{I-DML}) and the body branch (\emph{B-DML}) as well as their combination with each other (\emph{IB-DML}) and with the semantic segmentation branch (\emph{Sem-B-DML} and \emph{Sem-IB-DML}).
In the following, we also refer to our full model \emph{Sem-IB-DML} as \emph{Affect-DML}.

Our quantitative evaluation highlights the advantage of leveraging the  semantic segmentation-driven embedding  $f_{sem}$ both with and without the body-focused network.
Exchanging the body network $f_{body}$ from \emph{IB-DML} with our semantic network \emph{Sem-I-DML} provides a remarkable performance gain of $11.14\%$ on \emph{CAT-6:6} and about $5.29\%$ points on \emph{CAT-6:4}. 
Combining the representations of all three networks (\emph{Sem-IB-DML}) further increases the performance by $1.60\%$ on \emph{CAT-6:6} and $2.78\%$ on \emph{CAT-6:4}.
Our method also outperforms the original method of Kosti~\textit{et al.}~\cite{emotic_dataset_2} by a significant margin.

\subsection{One-shot Numerical Recognition Results}
\label{subsec:experiments:human_affect_levels}
Table~\ref{tab:experiment_level} lists the one-shot evaluation results of our  numerical human affect recognition benchmark on the two train-test splits indicated by \emph{LEV-7:3} and \emph{LEV-6:4} (see Section \ref{sec:problem}).
Since leveraging the body representation branch and the image feature extraction network outperforms Kosti~\textit{et al.}~\cite{emotic_dataset_2} extended with DML (\emph{IB-DML}$^{\red{\dagger}}$) in most metrics.

\noindent Once again, combining the full image representation branch with our semantic embedding branch $f_{sem}$ from \emph{SEM-I-DML} instead of a body feature extraction network network from \emph{IB-DML} improved the performance by $2.16\%$ for average accuracy on the three numerical human affect dimensions \emph{valence}, \emph{arousal} and \emph{dominance} (split \emph{LEV-7:3}).
Note, that we still evaluate the numerical emotion with categorical accuracy (assigning the predictions to ten partitions of the continuous space as the annotation in \cite{emotic_dataset}).
The gain in accuracy is consistent and also holds on the \emph{LEV-6:4} train-test split where \emph{SEM-I-DML} improves the  results by $3.30\%$ points over \emph{IB-DML}.
In contrast to the experiments on \emph{LEV-7:3}, combining all three networks yields a further  improvement of about $2.36\%$ compared with \emph{Sem-I-DML}, illustrating that the modalities mostly complement each other and their fusion is beneficial to our task.

\subsection{Architecture Analysis and Qualitative Results}
\label{subsec:experiments:qualitative}

In Table~\ref{tab:experiments_ablation}, we analyze the influence of the backbone architecture size on the recognition results (ResNet18 vs. ResNet50~\cite{ResNet}).
The ResNet50 architectures indicates better results on some tasks like \emph{CAT-6:6}, the recognition of \emph{valence} and \emph{arousal} on \emph{LEV-7:3} and the recognition of \emph{arousal} on \emph{LEV-6:4} while the ResNet18 architecture performs better on the other tasks.
When taking a closer look, the performance gains of ResNet50 are significant, while in cases where the smaller architecture shows advantages, the ResNet18 architecture only has slight performance gain of $0.4\%$ on average over where it outperforms ResNet50. 

\begin{figure}[t]
\begin{center}
\includegraphics[width=0.95\linewidth]{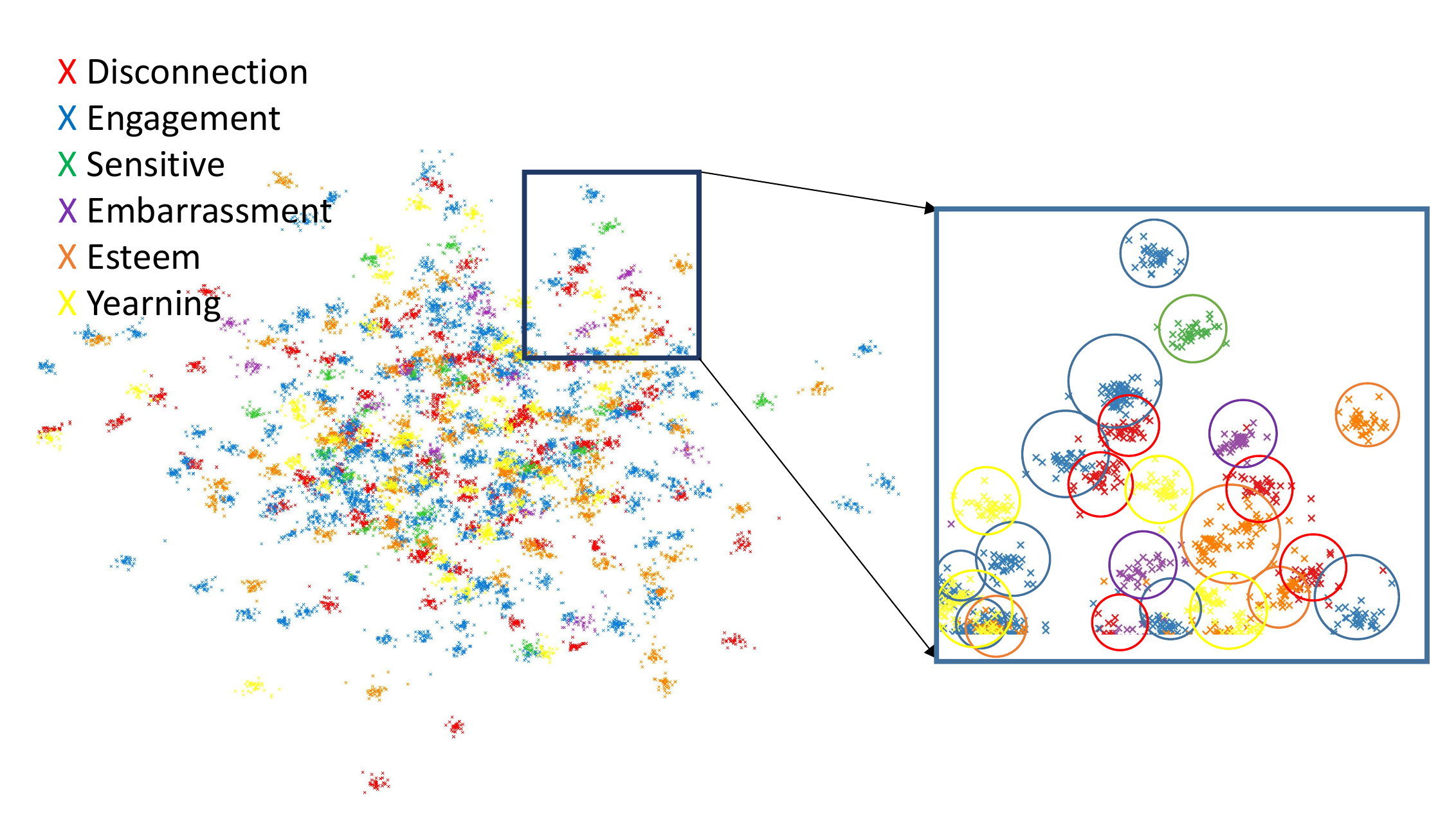}

\end{center}
\caption{A visualization of the intermediate embeddings for the unseen emotion classes as a 2-dimensional t-SNE representation. There is a clear separation of the individual clusters even though the categories were not present during training. Best viewed in color.}
\label{fig:tsne}
\end{figure}

In Figure~\ref{fig:tsne} we provide a 2-dimensional t-SNE~\cite{van2008visualizing} analysis of the latent space embedding learnt by  our best model \emph{Affect-DML}.
Each embedding has been colored according to its categories in testing set.
There is a clear separation of the individual clusters (the color represents the ground truth emotion category).
Note, that the visualization only covers categories not present during the classifier training.
This indicates, that  representation learned through combination of two loss functions provides representations which are highly discriminative and generalize well to novel classes, as the visualized states have not been seen during training.

Finally, we provide multiple qualitative prediction results in  Figure~\ref{fig:qualitative}. 
Our \emph{Affect-DML} model is referenced by [a] and compared with \emph{IB-DML$^{\red{\dagger}}$ } (Kosti~\textit{et al.}~\cite{emotic_dataset_2} extended with DML), referenced by [b].
In most examples, \emph{Affect-DML} provides better predictions, a result which is consistent with the quantitative evaluation in Table~\ref{tab:experiments_category} where \emph{Sem-IB-DML} (\emph{Affect-DML}) outperforms \emph{IB-DML} by $8.07\%$ for \emph{CAT-6:4}.

\section{CONCLUSION}
In this paper, we proposed \textit{Affect-DML} -- a framework for recognizing human emotions in context given a single visual example. 
Such one-shot recognition of human affect is a challenging but important task, because although the range of facial expressions remains the same, psychological categorization schemes and contexts-of-interest which  influence the emotions continuously change.
We formalize the one-shot categorization problem by augmenting the Emotic dataset and allowing only a set of coarse emotions be present during training, while during inference the model recognizes more fine-grained emotions from a single image example.
Our framework differs from the existing emotion categorization methods in two ways. First, we optimize our model using deep metric learning, leveraging a combination of cross entropy- and triplet loss. 
Second, we propose to leverage an encoding of the maps produced by a semantic segmentation network, which we refer to as  \textit{Sem2Vec}. 
We believe, that semantic segmentation information is vital for generalizable models of emotions \textit{in context}, which is consistently confirmed in our extensive experiments.

\mypar{Acknowledgments.} The research leading to these results was supported by the SmartAge
project sponsored by the Carl Zeiss Stiftung (P2019-01-003; 2021-2026).

{\small
\bibliographystyle{ieee}
\bibliography{egbib}

\begin{thebibliography}{10}\itemsep=-1pt

\bibitem{facial_depression_disease}
I.~M. Anderson et~al.
\newblock State-dependent alteration in face emotion recognition in depression.
\newblock {\em The British Journal of Psychiatry}, 2011.

\bibitem{antoniadis2021exploiting}
P.~Antoniadis, P.~P. Filntisis, and P.~Maragos.
\newblock Exploiting emotional dependencies with graph convolutional networks
  for facial expression recognition.
\newblock {\em arXiv}, 2021.

\bibitem{facial_parkinson_disease}
S.~Argaud, M.~V{\'e}rin, P.~Sauleau, and D.~Grandjean.
\newblock Facial emotion recognition in {Parkinson's} disease: {A} review and
  new hypotheses.
\newblock {\em Movement Disorders}, 2018.

\bibitem{triplet_margin_loss}
V.~Balntas, E.~Riba, D.~Ponsa, and K.~Mikolajczyk.
\newblock Learning local feature descriptors with triplets and shallow
  convolutional neural networks.
\newblock In {\em BMVC}, 2016.

\bibitem{fabian2016emotionet}
C.~F. Benitez{-}Quiroz, R.~Srinivasan, and A.~M. Mart{\'{\i}}nez.
\newblock {EmotioNet:} {An} accurate, real-time algorithm for the automatic
  annotation of a million facial expressions in the wild.
\newblock In {\em CVPR}, 2016.

\bibitem{carreira2017quo}
J.~Carreira and A.~Zisserman.
\newblock Quo vadis, action recognition? {A} new model and the kinetics
  dataset.
\newblock In {\em CVPR}, 2017.

\bibitem{psychological_27}
A.~S. Cowen and D.~Keltner.
\newblock Self-report captures 27 distinct categories of emotion bridged by
  continuous gradients.
\newblock {\em PNAS}, 2017.

\bibitem{cruz2014one}
A.~C. Cruz, B.~Bhanu, and N.~S. Thakoor.
\newblock One shot emotion scores for facial emotion recognition.
\newblock In {\em ICIP}, 2014.

\bibitem{imagenet}
J.~Deng, W.~Dong, R.~Socher, L.-J. Li, K.~Li, and L.~Fei-Fei.
\newblock {ImageNet:} {A} large-scale hierarchical image database.
\newblock In {\em CVPR}, 2009.

\bibitem{dosovitskiy2014discriminative}
A.~Dosovitskiy, J.~T. Springenberg, M.~Riedmiller, and T.~Brox.
\newblock Discriminative unsupervised feature learning with convolutional
  neural networks.
\newblock In {\em NeurIPS}, 2014.

\bibitem{du2014compound}
S.~Du, Y.~Tao, and A.~M. Martinez.
\newblock Compound facial expressions of emotion.
\newblock {\em PNAS}, 2014.

\bibitem{ekman1999basic}
P.~Ekman.
\newblock Basic emotions.
\newblock {\em Handbook of Cognition and Emotion}, 1999.

\bibitem{guo2017one}
Y.~Guo and L.~Zhang.
\newblock One-shot face recognition by promoting underrepresented classes.
\newblock {\em arXiv}, 2017.

\bibitem{ResNet}
K.~He, X.~Zhang, S.~Ren, and J.~Sun.
\newblock Deep residual learning for image recognition.
\newblock In {\em CVPR}, 2016.

\bibitem{facial_huitingtong_disease}
S.~M. Henley, M.~J. Novak, C.~Frost, J.~King, S.~J. Tabrizi, and J.~D. Warren.
\newblock Emotion recognition in {Huntington's} disease: {A} systematic review.
\newblock {\em Neuroscience \& Biobehavioral Reviews}, 2012.

\bibitem{koch2015siamese}
G.~Koch, R.~Zemel, and R.~Salakhutdinov.
\newblock Siamese neural networks for one-shot image recognition.
\newblock In {\em ICMLW}, 2015.

\bibitem{emotic_dataset}
R.~Kosti, J.~M. Alvarez, A.~Recasens, and A.~Lapedriza.
\newblock {EMOTIC:} {Emotions} in context dataset.
\newblock In {\em CVPRW}, 2017.

\bibitem{emotic_dataset_2}
R.~Kosti, J.~M. Alvarez, A.~Recasens, and A.~Lapedriza.
\newblock Context based emotion recognition using {EMOTIC} dataset.
\newblock {\em TPAMI}, 2020.

\bibitem{fe2003bayesian}
F.~Li, R.~Fergus, and P.~Perona.
\newblock A {Bayesian} approach to unsupervised one-shot learning of object
  categories.
\newblock In {\em ICCV}, 2003.

\bibitem{liu2017skeleton}
J.~Liu, A.~Shahroudy, D.~Xu, A.~C. Kot, and G.~Wang.
\newblock Skeleton-based action recognition using spatio-temporal {LSTM}
  network with trust gates.
\newblock {\em TPAMI}, 2018.

\bibitem{masi2018deep}
I.~Masi, Y.~Wu, T.~Hassner, and P.~Natarajan.
\newblock Deep face recognition: A survey.
\newblock In {\em SIBGRAPI}, 2018.

\bibitem{sl-dml}
R.~Memmesheimer, N.~Theisen, and D.~Paulus.
\newblock {SL-DML:} {Signal} level deep metric learning for multimodal one-shot
  action recognition.
\newblock In {\em ICPR}, 2021.

\bibitem{mittal2020emoticon}
T.~Mittal, P.~Guhan, U.~Bhattacharya, R.~Chandra, A.~Bera, and D.~Manocha.
\newblock {EmotiCon:} {Context-aware} multimodal emotion recognition using
  frege's principle.
\newblock In {\em CVPR}, 2020.

\bibitem{affectnet}
A.~Mollahosseini, B.~Hasani, and M.~H. Mahoor.
\newblock {AffectNet:} {A} database for facial expression, valence, and arousal
  computing in the wild.
\newblock {\em TAC}, 2019.

\bibitem{mou2015group}
W.~Mou, O.~Celiktutan, and H.~Gunes.
\newblock Group-level arousal and valence recognition in static images: Face,
  body and context.
\newblock In {\em FG}, 2015.

\bibitem{nicolaou2011continuous}
M.~A. Nicolaou, H.~Gunes, and M.~Pantic.
\newblock Continuous prediction of spontaneous affect from multiple cues and
  modalities in valence-arousal space.
\newblock {\em TAC}, 2011.

\bibitem{posner2005circumplex}
J.~Posner, J.~A. Russell, and B.~S. Peterson.
\newblock The circumplex model of affect: An integrative approach to affective
  neuroscience, cognitive development, and psychopathology.
\newblock {\em Development and Psychopathology}, 2005.

\bibitem{reiss2020activity}
S.~Rei{\ss}, A.~Roitberg, M.~Haurilet, and R.~Stiefelhagen.
\newblock Activity-aware attributes for zero-shot driver behavior recognition.
\newblock In {\em CVPRW}, 2020.

\bibitem{roitberg2020cnn}
A.~Roitberg, M.~Haurilet, S.~Rei{\ss}, and R.~Stiefelhagen.
\newblock {CNN-based} driver activity understanding: Shedding light on deep
  spatiotemporal representations.
\newblock In {\em ITSC}, 2020.

\bibitem{ruan2020context}
S.~Ruan et~al.
\newblock Context-aware generation-based net for multi-label visual emotion
  recognition.
\newblock In {\em ICME}, 2020.

\bibitem{emotion_healthy}
P.~Salovey, A.~J. Rothman, J.~B. Detweiler, and W.~T. Steward.
\newblock Emotional states and physical health.
\newblock {\em American Psychologist}, 2000.

\bibitem{santoro2016meta}
A.~Santoro, S.~Bartunov, M.~Botvinick, D.~Wierstra, and T.~Lillicrap.
\newblock Meta-learning with memory-augmented neural networks.
\newblock In {\em ICML}, 2016.

\bibitem{emotion_efficientnet}
A.~V. Savchenko.
\newblock Facial expression and attributes recognition based on multi-task
  learning of lightweight neural networks.
\newblock {\em arXiv}, 2021.

\bibitem{psychological_development}
K.~R. Scherer, V.~Shuman, J.~Fontaine, and C.~Soriano~Salinas.
\newblock The grid meets the wheel: Assessing emotional feeling via
  self-report.
\newblock {\em Components of Emotional Meaning: A Sourcebook}, 2013.

\bibitem{schindler2008recognizing}
K.~Schindler, L.~Van~Gool, and B.~De~Gelder.
\newblock Recognizing emotions expressed by body pose: A biologically inspired
  neural model.
\newblock {\em Neural Networks}, 2008.

\bibitem{shaban2017one}
A.~Shaban, S.~Bansal, Z.~Liu, I.~Essa, and B.~Boots.
\newblock One-shot learning for semantic segmentation.
\newblock In {\em BMVC}, 2017.

\bibitem{shaver1987emotion}
P.~Shaver, J.~Schwartz, D.~Kirson, and C.~O'connor.
\newblock Emotion knowledge: further exploration of a prototype approach.
\newblock {\em Journal of Personality and Social Psychology}, 1987.

\bibitem{emotion_review_physical}
L.~Shu et~al.
\newblock A review of emotion recognition using physiological signals.
\newblock {\em Sensors}, 2018.

\bibitem{soleymani2015analysis}
M.~Soleymani, S.~Asghari-Esfeden, Y.~Fu, and M.~Pantic.
\newblock Analysis of {EEG} signals and facial expressions for continuous
  emotion detection.
\newblock {\em TAC}, 2016.

\bibitem{facial_alzheimer_disease}
I.~Spoletini et~al.
\newblock Facial emotion recognition deficit in amnestic mild cognitive
  impairment and alzheimer disease.
\newblock {\em The American Journal of Geriatric Psychiatry}, 2008.

\bibitem{rmsprop}
T.~Tieleman and G.~Hinton.
\newblock {Lecture 6.5---RmsProp: Divide the gradient by a running average of
  its recent magnitude}.
\newblock COURSERA: Neural Networks for Machine Learning, 2012.

\bibitem{cnn_lstm_emotion}
P.~Tzirakis, G.~Trigeorgis, M.~A. Nicolaou, B.~W. Schuller, and S.~Zafeiriou.
\newblock End-to-end multimodal emotion recognition using deep neural networks.
\newblock {\em IEEE Journal of Selected Topics in Signal Processing}, 2017.

\bibitem{van2008visualizing}
L.~Van~der Maaten and G.~Hinton.
\newblock Visualizing data using {t-SNE}.
\newblock {\em JMLR}, 2008.

\bibitem{vinyals2016matching}
O.~Vinyals, C.~Blundell, T.~Lillicrap, K.~Kavukcuoglu, and D.~Wierstra.
\newblock Matching networks for one shot learning.
\newblock In {\em NeurIPS}, 2016.

\bibitem{HRNet}
J.~Wang et~al.
\newblock Deep high-resolution representation learning for visual recognition.
\newblock {\em TPAMI}, 2020.

\bibitem{wang2019robust}
L.~Wang, X.~Xu, F.~Liu, X.~Xing, B.~Cai, and W.~Lu.
\newblock Robust emotion navigation: Few-shot visual sentiment analysis by
  auxiliary noisy data.
\newblock In {\em ACIIW}, 2019.

\bibitem{multi_similarity}
X.~Wang, X.~Han, W.~Huang, D.~Dong, and M.~R. Scott.
\newblock Multi-similarity loss with general pair weighting for deep metric
  learning.
\newblock In {\em CVPR}, 2019.

\bibitem{wang2016learning}
Y.-X. Wang and M.~Hebert.
\newblock Learning to learn: Model regression networks for easy small sample
  learning.
\newblock In {\em ECCV}, 2016.

\bibitem{wolf2018estimating}
E.~Wolf, M.~Martinez, A.~Roitberg, R.~Stiefelhagen, and B.~Deml.
\newblock Estimating mental load in passive and active tasks from pupil and
  gaze changes using bayesian surprise.
\newblock In {\em ICMIW}, 2018.

\bibitem{zhan2019zero}
C.~Zhan, D.~She, S.~Zhao, M.-M. Cheng, and J.~Yang.
\newblock Zero-shot emotion recognition via affective structural embedding.
\newblock In {\em ICCV}, 2019.

\bibitem{places365}
B.~Zhou, A.~Lapedriza, A.~Khosla, A.~Oliva, and A.~Torralba.
\newblock Places: A 10 million image database for scene recognition.
\newblock {\em TPAMI}, 2018.

\bibitem{ade20k}
B.~Zhou, H.~Zhao, X.~Puig, S.~Fidler, A.~Barriuso, and A.~Torralba.
\newblock Scene parsing through {ADE20K} dataset.
\newblock In {\em CVPR}, 2017.

\end{thebibliography}
}

\end{document}